\documentclass[letterpaper, 10 pt, conference]{ieeeconf}  

\IEEEoverridecommandlockouts                              

\usepackage{amsmath,nccmath} 
\usepackage{amssymb}  
\usepackage{hyphenat}
\usepackage{graphicx}
\usepackage{balance}
\usepackage[hidelinks]{hyperref}
\usepackage[nonumberlist,acronym,toc,section,style=super,nopostdot,automake]{glossaries}
\makeglossaries
\usepackage{siunitx}
\usepackage{microtype}
\usepackage{nicefrac}
\usepackage{standalone}
\usepackage{subcaption}
\usepackage{bm} 
\usepackage{color}
\usepackage{tikz}
\usepackage{pgfplots}
\pgfplotsset{compat=newest}
\usetikzlibrary{shapes,animations}
\usetikzlibrary{matrix} 
\usetikzlibrary{arrows} 
\usetikzlibrary{arrows.meta} 
\usetikzlibrary{calc} 
\usetikzlibrary{decorations.markings}
\usetikzlibrary{backgrounds}
\usetikzlibrary{intersections}
\usetikzlibrary{fit}
\usetikzlibrary{positioning}
\usetikzlibrary{patterns}
\usepgfplotslibrary{groupplots}
\usepackage{bm}
\usepackage[outline]{contour}

\usepackage{hyperref}
\usepackage[normalem]{ulem}
\usepackage{float}

\title{\LARGE \bf
Next\hyp{}Best\hyp{}Trajectory Planning of Robot Manipulators for Effective Observation and Exploration}

\author{Heiko Renz, Maximilian Kr{\"a}mer, Frank Hoffmann and Torsten Bertram
\thanks{This work is funded by the Deutsche Forschungsgemeinschaft (DFG,
German Research Foundation) - 497071854.}
\thanks{All authors are with the Institute of Control Theory and Systems
Engineering, TU Dortmund University, Dortmund, Germany. {\tt\small [heiko.renz, maximilian.kraemer, frank.hoffmann, torsten.bertram]@tu-dortmund.de}}%
}

\makeatletter
\apptocmd{\@maketitle}{
    \centering
    \includegraphics[trim = 0cm 0cm 0cm 0cm, clip,width=0.24\linewidth]{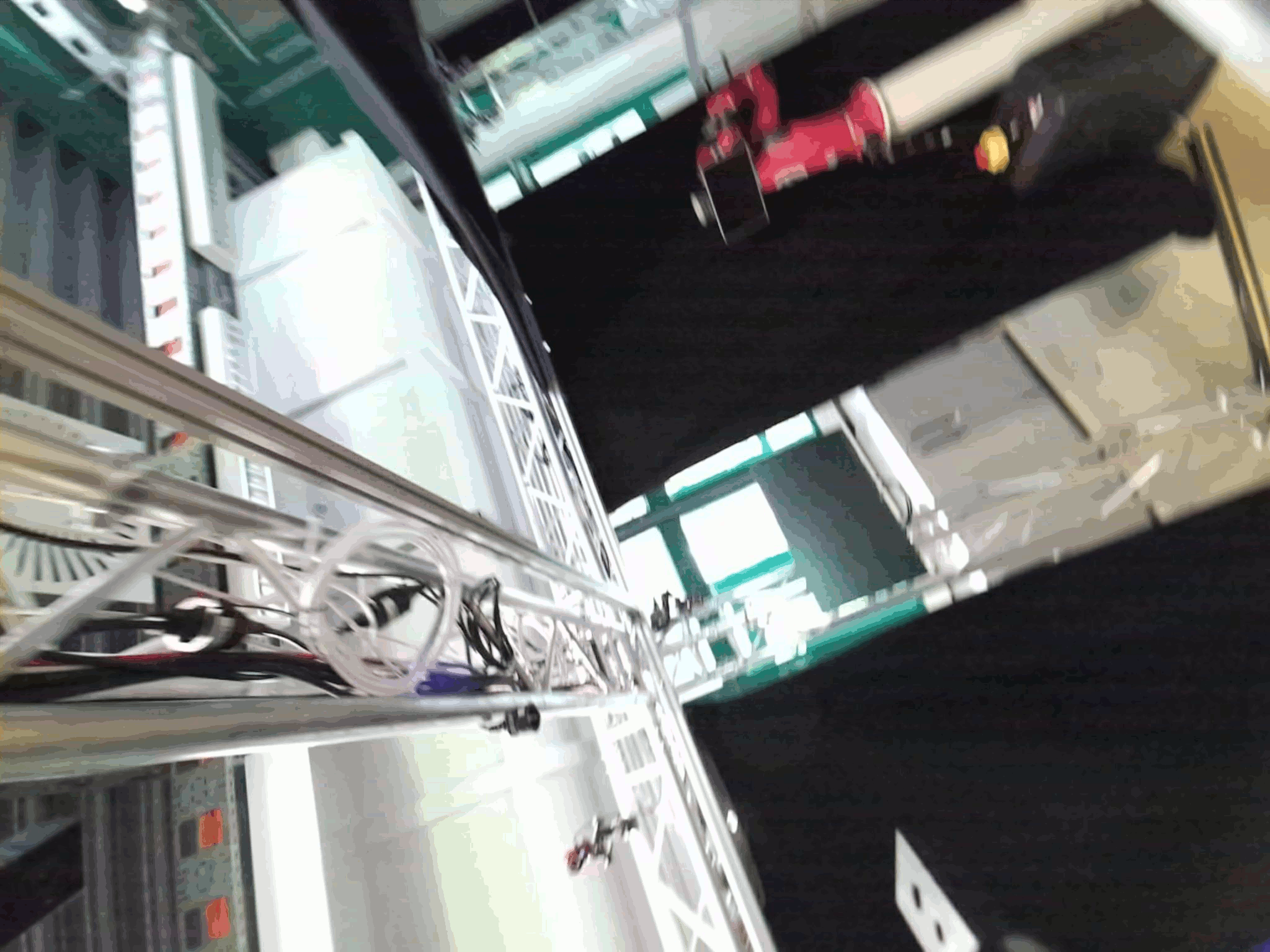}
    \includegraphics[trim = 0cm 0cm 0cm 0cm, clip,width=0.24\linewidth]{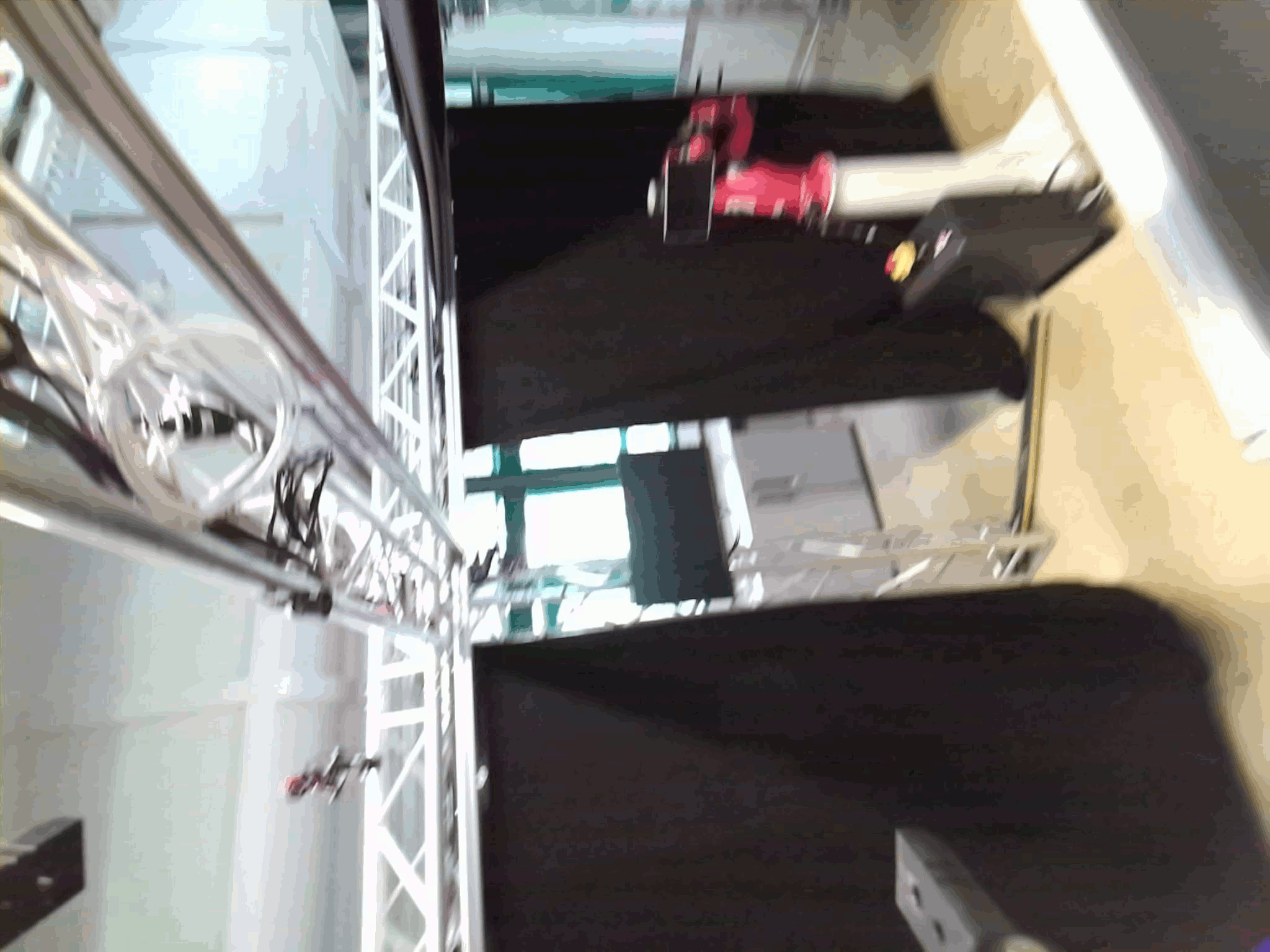}
    \includegraphics[trim = 0cm 0cm 0cm 0cm, clip,width=0.24\linewidth]{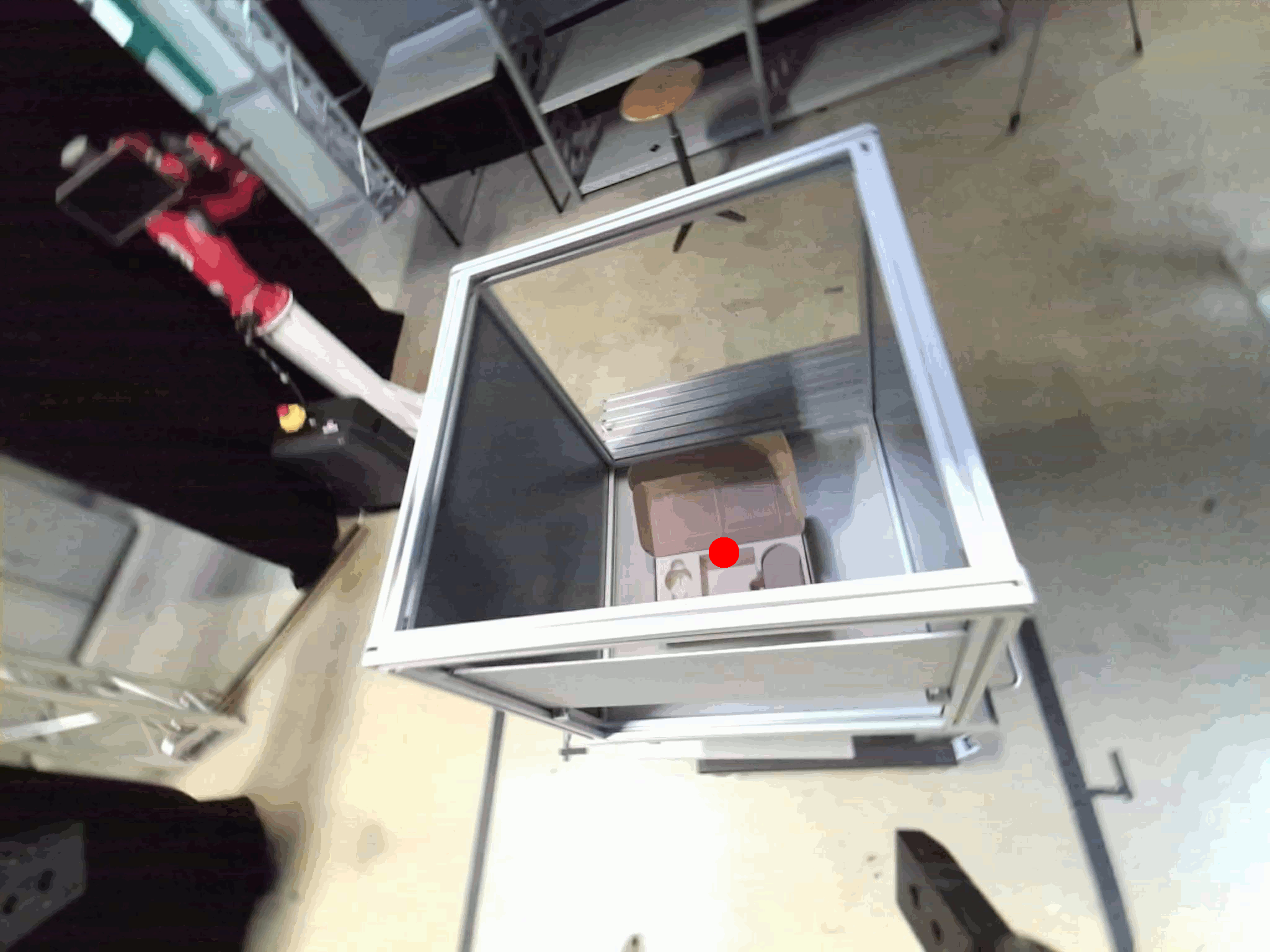}
    \includegraphics[trim = 0cm 0cm 0cm 0cm, clip,width=0.24\linewidth]{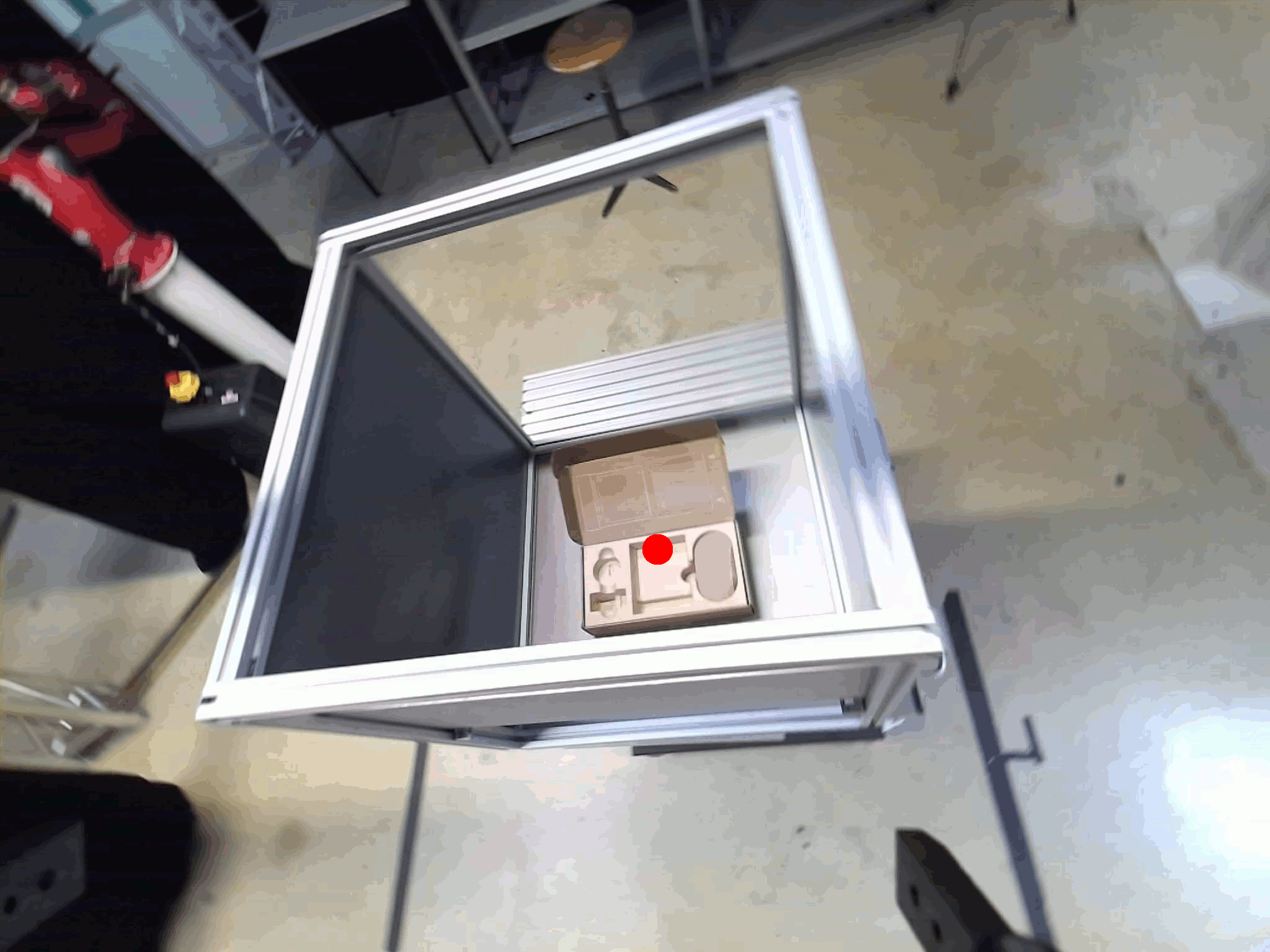}
    \hspace{\textwidth}

    \vspace{-1em}
    \includegraphics[width=0.24\linewidth]{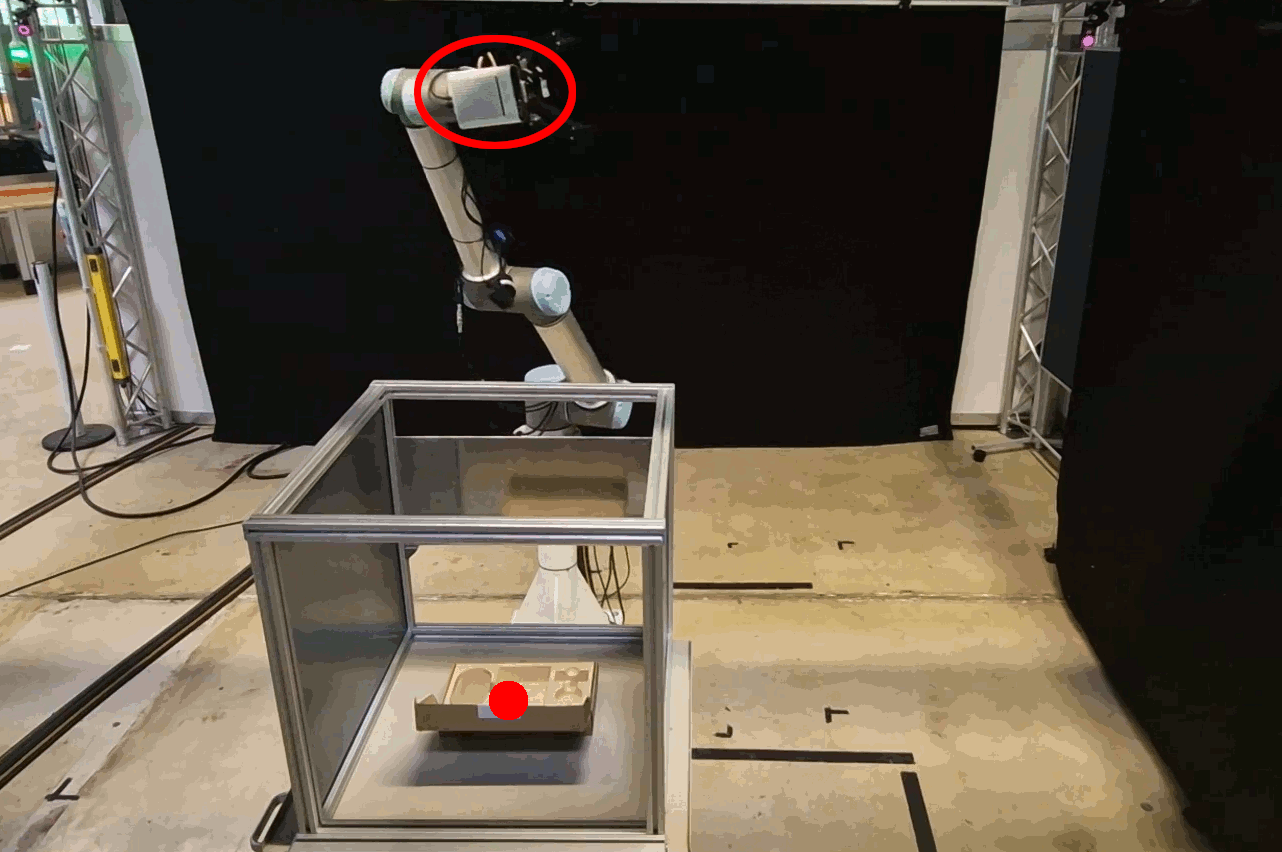}
    \includegraphics[
    width=0.24\linewidth]{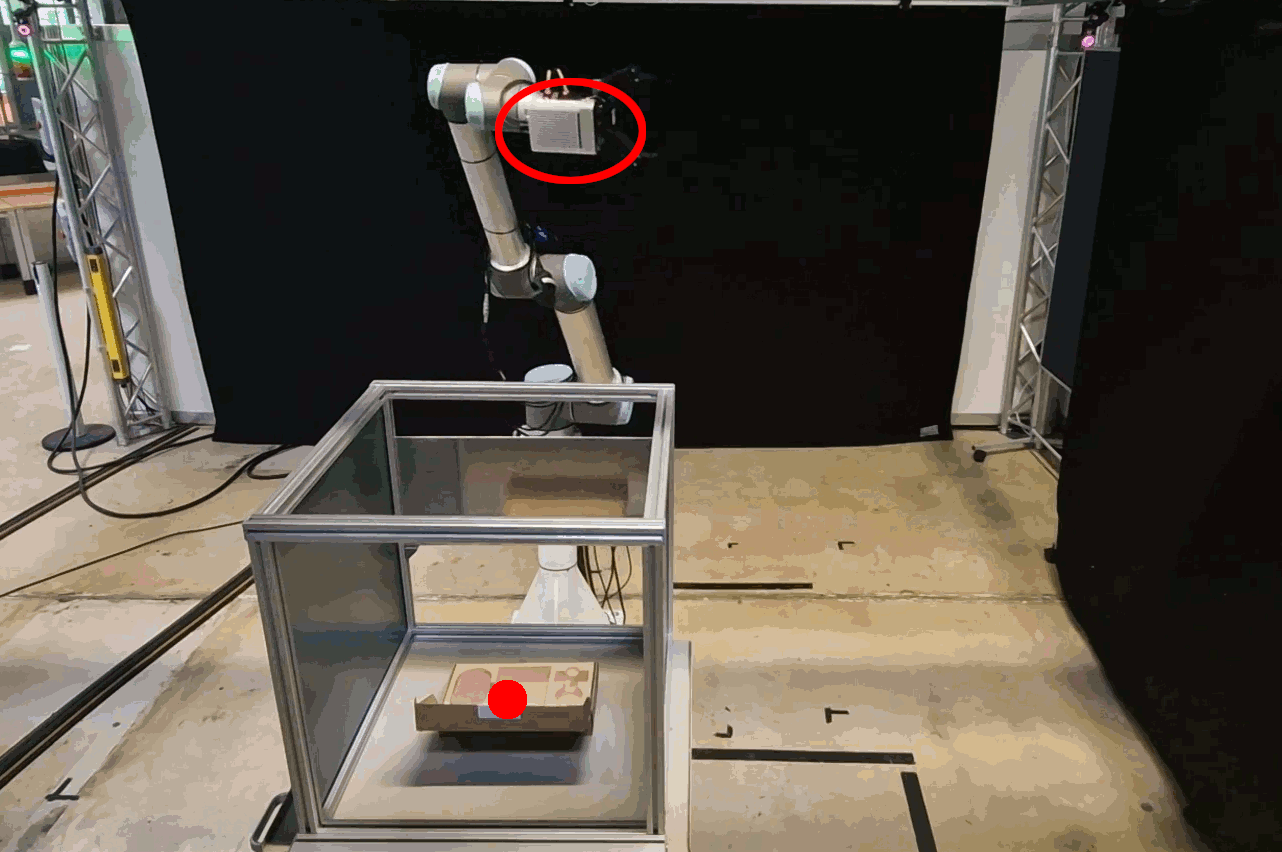}
    \includegraphics[width=0.24\linewidth]{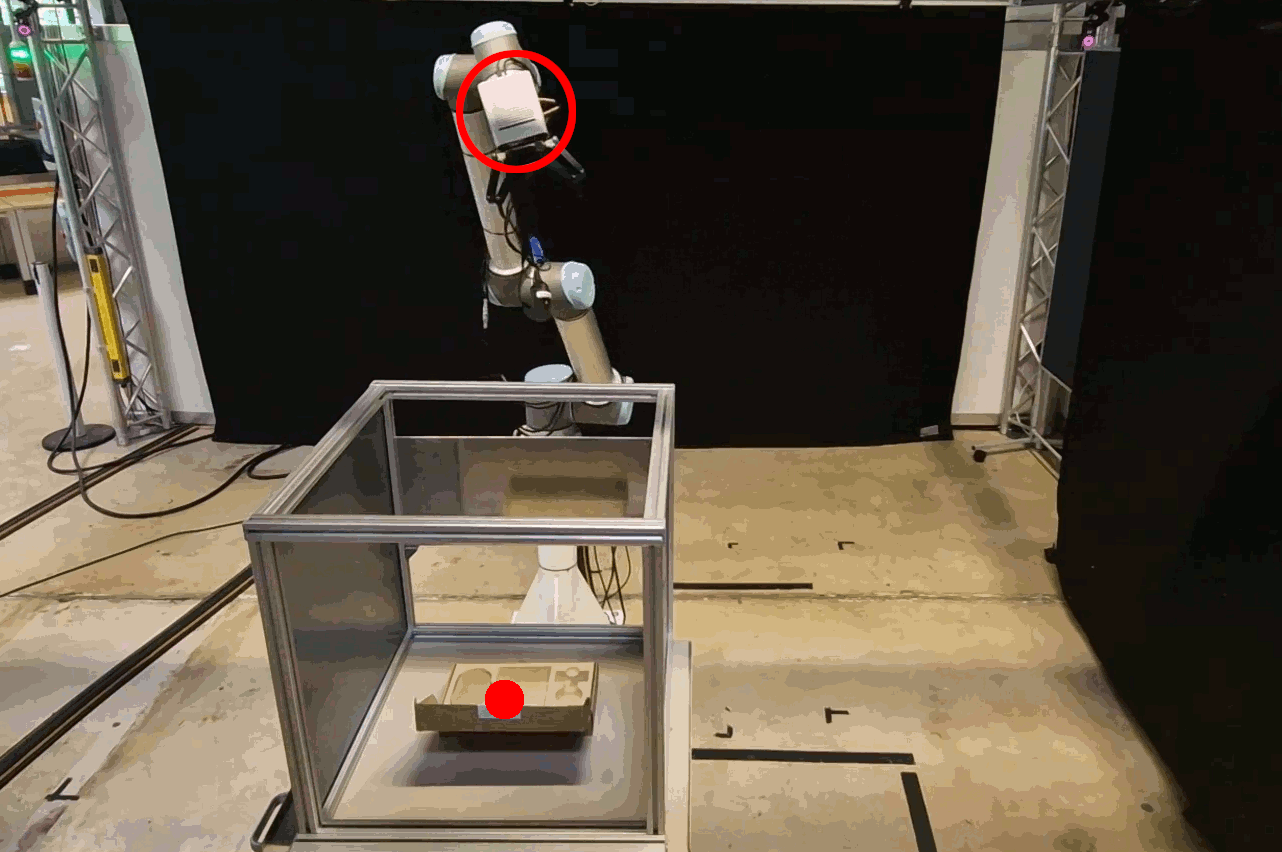}
    \includegraphics[width=0.24\linewidth]{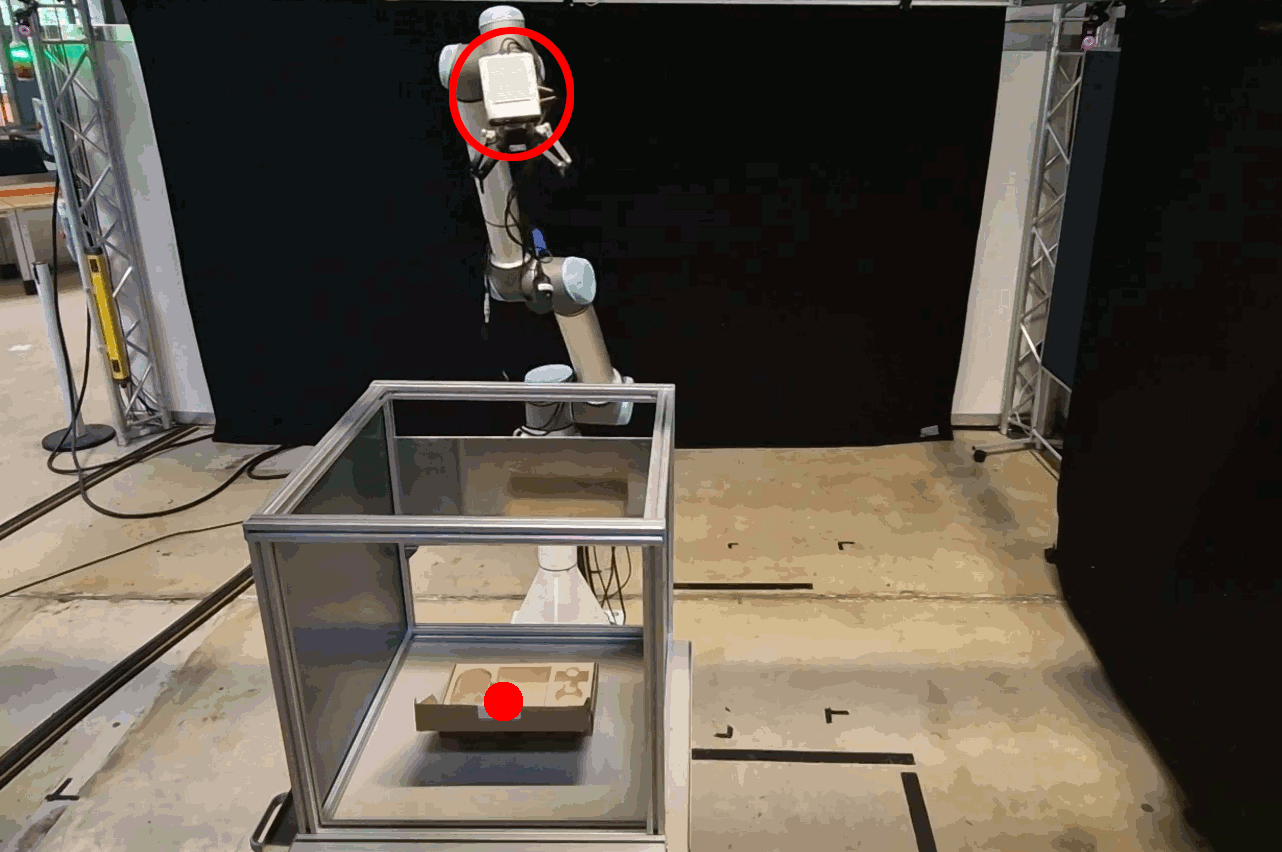}
    \hspace{\textwidth}
    \vspace{-1em}
    \captionof{figure}{The camera view (top row) and the robot (bottom row) without optimizing the Information Gain towards the Point of Interest (red dot) and during Next\hyp{}Best\hyp{}Trajectories with weights of $w_{\text{I}} = 5$, $w_{\text{I}} = 25$, and $w_{\text{I}} = 50$ (left to right).}
    \label{fig:teaser}
    \vspace{-1em}
   }{}{}
\makeatother

\newcommand\copyrighttext{%
    \footnotesize \textcopyright 2025 IEEE. Personal use of this material is permitted.
    Permission from IEEE must be obtained for all other uses, in any current or future
    media, including reprinting/republishing this material for advertising or promotional
    purposes, creating new collective works, for resale or redistribution to servers or
    lists, or reuse of any copyrighted component of this work in other works.}
\newcommand\copyrightnotice{%
    \begin{tikzpicture}[remember picture,overlay]
    \node[anchor=south,xshift=6pt,yshift=10pt] at (current page.south) {\fbox{\parbox{\dimexpr\textwidth-\fboxsep-\fboxrule\relax}{\copyrighttext}}};
    \end{tikzpicture}%
}

\begin{document}

\maketitle

\setcounter{figure}{1}
\thispagestyle{empty}
\pagestyle{empty}
\copyrightnotice

\newacronym{poi}{\ensuremath{\text{PoI}}}{\ensuremath{\text{Point of Interest}}}
\newacronym{gpu}{\ensuremath{\text{GPU}}}{\ensuremath{\text{Graphics Processing Unit}}}
\newacronym{cpu}{\ensuremath{\text{CPU}}}{\ensuremath{\text{Central Processing Unit}}}
\newacronym{fov}{\ensuremath{\text{FoV}}}{\ensuremath{\text{Field of View}}}
\newacronym{nlp}{\ensuremath{\text{NLP}}}{\ensuremath{\text{Nonlinear Programming}}}
\newacronym{idw}{\ensuremath{\text{IDW}}}{\ensuremath{\text{Inverse Distance Weighting}}}
\newacronym{id}{\ensuremath{\text{ID}}}{\ensuremath{\text{Information Distribution}}}
\newacronym[longplural={\ensuremath{\text{Next\hyp{}Best\hyp{}Trajectories}}}]{nbt}{\ensuremath{\text{NBT}}}{\ensuremath{\text{Next\hyp{}Best\hyp{}Trajectory}}}
\newacronym{nbv}{\ensuremath{\text{NBV}}}{\ensuremath{\text{Next\hyp{}Best\hyp{}View}}}
\newacronym{ig}{\ensuremath{\text{IG}}}{\ensuremath{\text{Information Gain}}}
\newacronym{auc}{\ensuremath{\text{AUC}}}{\ensuremath{\text{Area Under the Curve}}}
\newacronym{mhp}{\ensuremath{\text{MHP}}}{\ensuremath{\text{Moving Horizon Planner}}}
\newacronym{of}{\ensuremath{\text{OF}}}{\ensuremath{\text{Orientation Factor}}}
\newacronym{mpc}{\ensuremath{\text{MPC}}}{\ensuremath{\text{Model Predictive Control}}}
\newacronym{pcl}{\ensuremath{\text{PC}}}{\ensuremath{\text{Point Cloud}}}


\begin{abstract}

Visual observation of objects is essential for many robotic applications, such as object reconstruction and manipulation, navigation, and scene understanding.
Machine learning algorithms constitute the state\hyp{}of\hyp{}the\hyp{}art in many fields but require vast data sets, which are costly and time\hyp{}intensive to collect.
Automated strategies for observation and exploration are crucial to enhance the efficiency of data gathering.
Therefore, a novel strategy utilizing the \acrlong*{nbt} principle is developed for a robot manipulator operating in dynamic environments. 
Local trajectories are generated to maximize the information gained from observations along the path while avoiding collisions.
We employ a voxel map for environment modeling and utilize raycasting from perspectives around a point of interest to estimate the information gain.
A global ergodic trajectory planner provides an optional reference trajectory to the local planner, improving exploration and helping to avoid local minima.
To enhance computational efficiency, raycasting for estimating the information gain in the environment is executed in parallel on the graphics processing unit. 
Benchmark results confirm the efficiency of the parallelization, while real\hyp{}world experiments demonstrate the strategy's effectiveness.
\end{abstract}
\glslocalresetall
\section{Introduction}



Observing objects and exploring unknown environments are essential for robots to gather task\hyp{}relevant information. Observation collects object data, while exploration navigates uncharted areas for new insights.

This study presents a novel local trajectory planner that maximizes the \gls*{ig} along its path using a \gls*{gpu}-accelerated online local \gls*{id}. 
By aggregating \glspl*{ig} from multiple perspectives, the \gls*{id} offers to optimize the camera view during trajectory planning. 
A global ergodic trajectory planner \cite{dongTimeOptimalErgodic2023a} provides a reference trajectory, integrating exploration and observation while avoiding local minima.

The observation data, collected from various sensor modalities, models the environment with an online voxel map for the \gls*{nbt} strategy. 
Unlike \gls*{nbv} approaches, \gls*{nbt} maximizes the \gls*{ig} along the entire trajectory rather than focusing on a single perspective, offering a more comprehensive view. 
The voxel map estimates the \gls*{ig} from different perspectives, with unknown-state voxels holding the highest \gls*{ig} potential, while free or occupied voxels still provide useful environmental updates. 
A parallelized raycasting algorithm computes \glspl*{ig} within the environment. 
Fig.~\ref{fig:teaser} illustrates the benefits of prioritizing the \gls*{ig} when observing a \gls*{poi}.

The key contributions of this work are summarized as follows, and the code is publicly available as a branch of the \gls*{mhp}\footnote{\url{https://github.com/rst-tu-dortmund/mhp4hri/tree/next-best-trajectory-github}} \cite{renzMovingHorizonPlanning2024}:
\begin{itemize}
\item 
Online local \gls*{id} using raycasting to estimate \glspl*{ig} from multiple perspectives in a voxelized environment, building an \gls*{ig} \gls*{pcl} based on the voxel map;
\item Parallelization of the \gls*{ig} algorithm on a \gls*{gpu} to improve computational efficiency;
\item Integration of the \gls*{id} into an \gls*{mhp} to determine the \gls*{nbt} for a robot manipulator;
\item Extension of the \gls*{mhp} to track a global ergodic trajectory \cite{dongTimeOptimalErgodic2023a}, enhancing exploration and preventing standstills.
\end{itemize}



\section{Related Work}

This section explores mapping techniques for calculating \glspl*{ig} and robotic planning strategies integrating perspective information for goal selection.

\paragraph*{Mapping Techniques}
Mapping is fundamental in robotics for environmental representation and planning. The choice of technique \textendash{} ranging from \num{2.5}D height maps \cite{denglerViewpointPushPlanning2023} to voxel maps \cite{hornungOctoMapEfficientProbabilistic2013,minOctoMapRTFastProbabilistic2023,dubergUFOMapEfficientProbabilistic2020} and direct \gls*{pcl} analysis \cite{monicaContourbasedNextbestView2018} \textendash{} depends on specific needs and constraints.

GPU acceleration reduces the computational cost of 3D map generation from \glspl*{pcl} \cite{minOctoMapRTFastProbabilistic2023}. Voxel maps, notably \textit{OctoMap} \cite{hornungOctoMapEfficientProbabilistic2013}, use an octree structure to improve memory efficiency and address height map limitations. \textit{OctoMap\hyp{}RT} \cite{minOctoMapRTFastProbabilistic2023} extends this by leveraging \gls*{gpu} for faster map generation.
Other GPU-based voxel mapping methods include \textit{G\hyp{}VOM} \cite{overbyeGVOMGPUAccelerated2022a}, optimized for real-time off-road exploration, and \textit{Voxblox++} \cite{grinvaldVolumetricInstanceAwareSemantic2019}, which integrates RGB-D data for instance\hyp{}aware semantic mapping. 
\textit{UfoMap} \cite{dubergUFOMapEfficientProbabilistic2020} improves \textit{OctoMap} by explicitly marking free, occupied, and unknown areas, making it ideal for \gls*{ig} tasks. We use \textit{UfoMap} for its ability to represent unknown space and its publicly available code, setting it apart from alternatives.



\paragraph*{Reconstruction and Exploration}

Many mapping and robotic exploration strategies survey or reconstruct environments and objects. Robots navigate different locales to acquire new insights, strategizing their motion toward the optimal perspective \textendash{} the \gls*{nbv} \textendash{} to enhance environmental knowledge \cite{denglerViewpointPushPlanning2023,bircherRecedingHorizonNextBestView2016,luOptimalFrontierBasedAutonomous2020,santosAutonomousSceneExploration2020}.

One approach to explore unknown areas employs drones with sampling-based planners \cite{bircherRecedingHorizonNextBestView2016}. 
Frontier-based strategies further improve exploration efficiency \cite{luOptimalFrontierBasedAutonomous2020}. 
Research also extends to robotic manipulators, with voxel-based methods enabling autonomous scene navigation and analysis \cite{santosAutonomousSceneExploration2020}.
A raycasting algorithm evaluates potential \glspl*{ig} from randomized perspectives around the \gls*{poi} to estimate the \gls*{nbv}. 
However, it fails to account for occlusions or misorientation as the robot moves. 
Increasing perspectives mitigates occlusions but requires constant replanning. 
The \gls*{nbt} strategy conversely optimizes the entire trajectory to maximize gain under given constraints instead of individual perspectives.
Direct robot-environment interaction is crucial for capturing details in unknown environments \cite{denglerViewpointPushPlanning2023}, particularly for revealing hidden objects, though it is less effective for documenting changes.

Object reconstruction is another key task where \glspl*{nbv} provide guidance. 
Incorporating uncertainties and probabilities into \gls*{ig} calculations improves reconstruction quality \cite{delmericoComparisonVolumetricInformation2018}. 
Expanding the \gls*{nbv} approach to sequence perspectives further enhances reconstruction \cite{panGlobalGeneralizedMaximum2023}.

\paragraph*{\acrlong*{nbt}}
However, a limitation of these methods is their failure to account for the potential \glspl*{ig} that could arise between perspectives despite optimizing sequences for reconstruction.
An \gls*{nbt} approach has been proposed for drone exploration to optimize the drone's trajectory in a tree\hyp{}based manner, considering possible \glspl*{ig} in a voxel map \cite{lindqvistTreeBasedNextBestTrajectoryMethod2024}.
The method also applies \textit{UfoMap} due to modeling unknown space and embeds the \gls*{ig} into an exploration-rapidly-exploring-random tree and a nonlinear model predictive control framework to control the drone.
Further work includes ergodic coverage and optimization techniques \cite{dongTimeOptimalErgodic2023a}.
These methods allow exploration based on ergodicity, ensuring that the robot explores the environment uniformly and efficiently or based on a desired target distribution representing areas of different estimated \glspl*{ig}.
These methods, however, are designed and tested on mobile robots or drones and not for robot manipulators in shared workspaces with humans or are not executable online.
Nevertheless, an adapted version of \cite{dongTimeOptimalErgodic2023a} extends our local trajectory planner to follow a global ergodic trajectory, exploring the environment and avoiding freezing robot problems.

\section{Online Local \acrlong*{id}}
\label{sec:infoDist}

Our approach extends \textit{UfoMap} \cite{dubergUFOMapEfficientProbabilistic2020} to estimate \glspl*{ig} from various perspectives around a \gls*{poi} whenever a new \gls*{pcl} is captured.
New \glspl*{pcl} are captured online during robot motion with an end\hyp{}effector camera, accounting for new occlusions and environmental changes.
These perspective-based \glspl*{ig} contribute to building a local \gls*{id}, enabling the evaluation of \gls*{ig} for the \gls*{nbt} at different trajectory poses. 
The local \gls*{id} is linked to entropy, which measures the uncertainty in voxel occupancy.
High entropy signifies a high level of uncertainty, which corresponds to a high value in the \gls*{id}. 
This is because perspectives that reduce environmental uncertainty result in a significant increase in \gls*{ig}.

The first step of generating the \gls*{id} applies a voxel filter from the \textit{Point Cloud Library} \cite{rusu3DHerePoint2011} to thin out the \gls*{pcl}, thus boosting the voxel map creation. 
Generating the \gls*{id} encompasses three main steps: selecting perspectives, calculating endpoints, and determining \gls*{ig} through raycasting.

\paragraph*{Perspective Selection}
To establish a local \gls*{id}, different starting points for perspectives $\mathbf{p}_{\text{P}} = {\left[p_{\text{P,x}}~p_{\text{P,y}}~p_{\text{P,z}}\right]}^{\text{T}} \in \mathbb{R}^{3}$ are sampled around the chosen \gls*{poi} $\mathbf{p}_{\text{POI}} = {\left[p_{\text{POI,x}}~p_{\text{POI,y}}~p_{\text{POI,z}}\right]}^{\text{T}} \in \mathbb{R}^{3}$.
The \gls*{poi} is modeled as a single point in the center of an object, simplifying the process of calculating the \gls*{id}.
A uniform distribution of $N_{\text{P}}$ perspectives within a spherical area surrounding the \gls*{poi} is adopted for thorough coverage. 
To ensure an even spread, observation points are generated on the sphere's surface, done efficiently using Muller's method \cite{mullerNoteMethodGenerating1959}.
The points initially generated on the unit sphere's surface are scaled to match the desired sphere radius $r_{\text{S}}$ around the \gls*{poi}. 
To achieve a uniform distribution of points in this sphere, the surface points are scaled by a uniformly distributed value $X_{\text{R}} \sim \mathcal{G}(0,1)$ as: 
\begin{equation}
\mathbf{p}_{\text{P}} = r_{\text{S}} \: X_{\text{R}}^{\nicefrac{1}{3}} \: \frac{X_k }{\sqrt{\sum_{l\in \{x,y,z\}} X_l^{2} }} \quad \text{for} \quad k \in \{x,y,z\},
\end{equation} 
with normal distributed values $X_{\{x,y,z\}} \sim \mathcal{N}(\mu = 0, \sigma = 1)$ in the Euclidean space $x,y,z$.
Incorporating the $\nicefrac{1}{3}$ power is required for achieving a uniform distribution of points within the sphere, accounting for the 3D sphere's surface area variation. 
For the radius $r_{\text{S}}$, a value of $r_{\text{S}} = \SI{1}{\m}$ is chosen to ensure that the perspectives are close enough to the \gls*{poi} to capture details.
The orientation of each perspective is then determined by the vector originating from the perspective's starting point $\mathbf{p}_{\text{P}}$ to the \gls*{poi} $\mathbf{p}_{\text{POI}}$

\begin{figure}[tb]
    \centering
    \vspace{5pt} 
    \begin{subfigure}{0.48\columnwidth}
        \centering
        \includegraphics[width=\columnwidth]{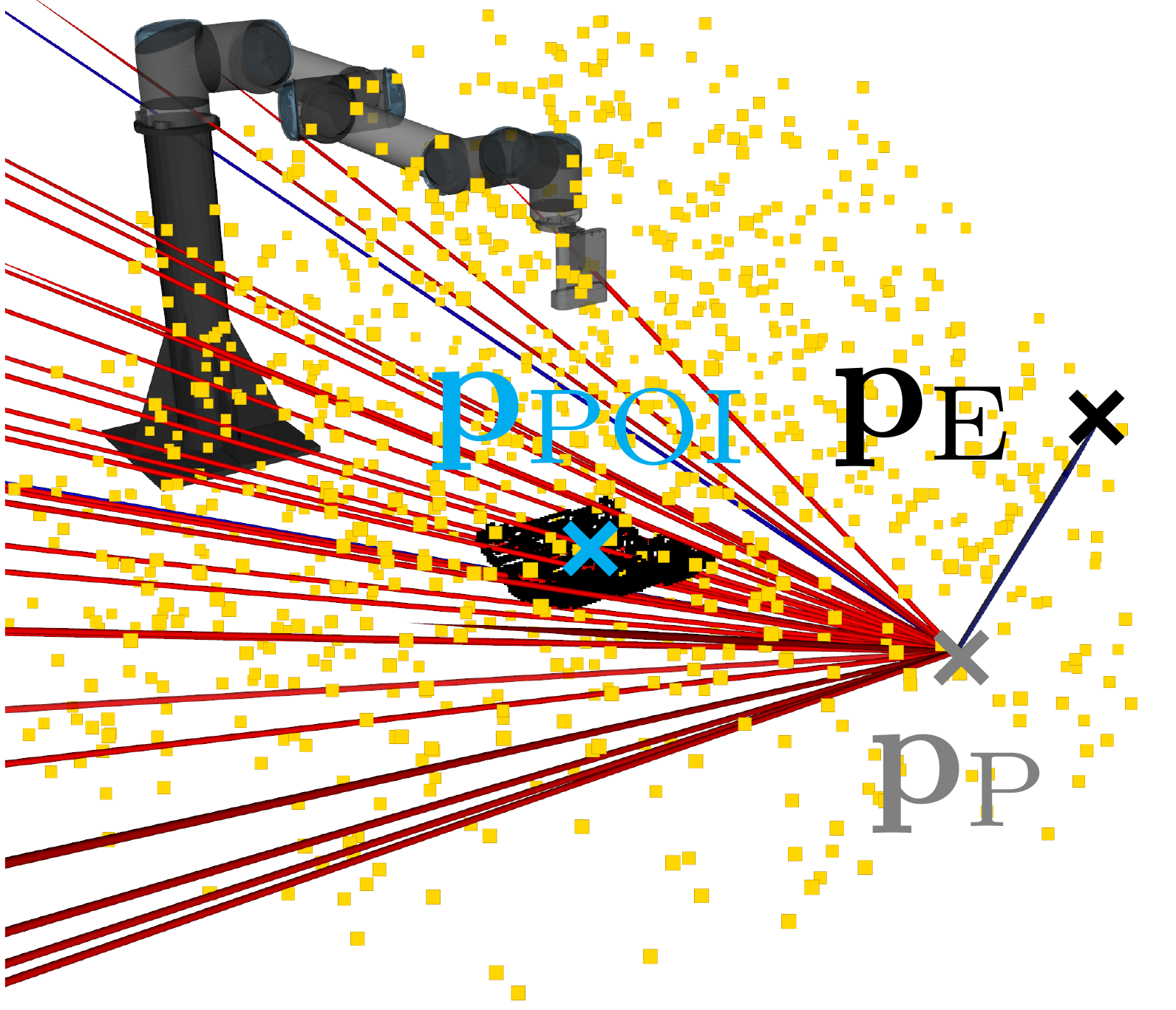}
        \captionsetup{belowskip=-5pt,aboveskip=-0pt}
        \caption{Perspectives (yelllow) including exemplary rays for a single perspective (red, corners in blue) with $s_{\text{G}} = 100$\\}
        \label{fig:raycasts}
    \end{subfigure}%
    \hfill
    \begin{subfigure}{0.48\columnwidth}
        \centering
        \includegraphics{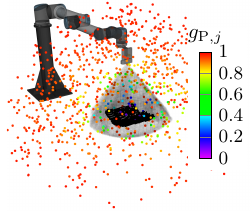}
        \captionsetup{belowskip=-5pt,aboveskip=-0pt}
        \caption{Local \gls*{id} with $s_{\text{G}} = 50$. Color representing the \gls*{ig} for each perspective. The grey cone marks free voxels in front of objects in the camera \gls*{fov}.
        }
        \label{fig:infoDist}  
    \end{subfigure}
    \captionsetup{belowskip=-15pt}
    \caption{Raycasting (\subref{fig:raycasts}) and \gls*{id} results (\subref{fig:infoDist}) with $N_{\text{P}} = 1000$.}
    \label{fig:combined_raycasts}
\end{figure}

\paragraph*{Endpoint Calculation}
Once the perspectives $\mathbf{p}_{\text{P}}$ are determined, the process calculates for each $\mathbf{p}_{\text{P}}$ $N_{\text{E}}$ endpoints $\mathbf{p}_{\text{E}} = {\left[p_{\text{E,x}}~p_{\text{E,y}}~p_{\text{E,z}}\right]}^{\text{T}} \in \mathbb{R}^{3}$ inside the camera's \gls*{fov}. 
Each perspective's frustum center is oriented towards the \gls*{poi}, and the camera's far frustum plane is at the camera's range $d_{\text{Cam}}$.
The central point 
is adjusted along the $x$ and $y$ axes by $d_{\text{h}}$ and $d_{\text{v}}$ to calculate the peripheral points on the camera's far frustum plane. 
This adjustment considers the camera's vertical and horizontal \gls*{fov} ($\text{\gls*{fov}}_{\text{v}}$ and $\text{\gls*{fov}}_{\text{h}}$) and its range $d_{\text{Cam}}$ using the formulas $d_{\text{h}} = d_{\text{Cam}} \cdot \tan\left(\nicefrac{\text{\gls*{fov}}_{\text{h}}}{2}\right)$ and $d_{\text{v}} = d_{\text{Cam}} \cdot \tan\left(\nicefrac{\text{\gls*{fov}}_{\text{v}}}{2}\right)$.
The calculated points delineate the four border rays of the frustum and serve as limitations for the endpoint rays.
Evaluating \glspl*{ig} from only the frustum endpoints is insufficient since, depending on the camera's parameters $d_{\text{Cam}}$, $\text{\gls*{fov}}_{\text{v}}$ and $\text{\gls*{fov}}_{\text{h}}$, these border rays exhibit distances that can be greater than objects to model.
Hence, $N_{\text{E}}$ endpoints $\mathbf{p}_{\text{E}}$ are sampled within the frustum's borders on a grid. 
The distance between these endpoints is determined by the voxel resolution $s_{\text{Vox}}$ set in the \textit{UfoMap} implementation, facilitating a calculation of \gls*{ig} at the finest level of environmental detail the map can represent. 
However, a challenge arises from the sheer number of endpoints and the computational demands of raycasting, especially when $s_{\text{Vox}}$ is small. 
To mitigate this, a variable grid size scaling factor $s_{\text{G}}$ is introduced, allowing the adjustment of endpoint density relative to $s_{\text{Vox}}$, effectively balancing accuracy with computational efficiency. 
Fig.~\ref{fig:raycasts} illustrates raycasting for a scaling factor $s_{\text{G}}=100$ and a fixed number of perspectives $N_{\text{P}} = 1000$, including introduced points.

\paragraph*{\gls*{ig} Calculation}

To compute the \gls*{ig} from each perspective $\mathbf{p}_{\text{P}}$, raycasting is conducted from every $\mathbf{p}_{\text{P}}$ to all related $N_{\text{E}}$ endpoints $\mathbf{p}_{\text{E}}$, evaluating the voxel states along each ray. 
The calculation employs a straightforward metric 
for each perspective  $j = 0,1,\dots,N_{\text{P}}-1$, based on the voxel's state and probability. 
This metric is designed for rapid computation, essential for real-time processing across numerous perspectives and endpoints. 
The \gls*{ig} $g_{\text{R},k, j}$ for each ray $k = 0,1,\dots,N_{\text{E}}-1$ of perspective $j$, takes states $v_i \in \{\textit{Occupied},\textit{Free},\textit{Unknown}\}$ of all encountered voxels $N_{\text{V}}$ on the ray into account. 
It reflects a cumulative assessment  $g_{\text{R},k, j} = \sum_{i=0}^{N_{\text{O}}} g(v_i)$ of \glspl*{ig} $g(v_i)$ based on whether voxels $v_i$ are occupied, free, or unknown, as classified by the \textit{UfoMap} based on occupancy probabilities $P\left(v_i\right)$:
\begin{equation}
    g(v_i) = \begin{cases}
        1-P(v_i) & \text{if } v_i = \textit{Occupied},\\
        P(v_i) & \text{if } v_i = \textit{Free},\\
        1 & \text{if } v_i = \textit{Unknown}.
        \end{cases}
\end{equation}
$N_{\text{O}}$ represents the count of voxels encountered on a ray from perspective $\mathbf{p}_{\text{P}}$ to the first occupied voxel or the endpoint, whichever comes first; hence, $N_{\text{O}} \leq N_{\text{V}}$.
The average \gls*{ig} $g_{\text{P},j}$ for a perspective $j$, denoted as $g_{\text{P},j}$, is calculated by averaging the \glspl*{ig} from all rays emanating from that perspective, formulated as $g_{\text{P},j} = \frac{1}{N_{\text{E}}} \sum_{k=0}^{N_{\text{E}}-1} g_{\text{R},k, j}$.

Fig.~\ref{fig:infoDist} illustrates a local \gls*{id} for $N_{\text{P}} = 1000$ perspective \glspl*{ig} with a grid scaling factor of $s_{\text{G}} = 50$.
This visualization highlights that perspectives nearer to the \gls*{poi}, particularly within the object's vicinity, tend to yield lower \glspl*{ig} because rays starting closer to the \gls*{poi} encounter the object sooner. 

Given its iterative nature over numerous perspectives, endpoints, and voxels for each ray, calculating this \gls*{ig} is highly computationally intensive. 
This process is suited for parallel processing on a \gls*{gpu} to reduce computation time significantly. 
A CUDA-based implementation of this algorithm facilitates the parallel evaluation of a vast number of perspectives and rays. 
This implementation is available in the code repository, and the performance is assessed in Section~\ref{sec:experiments}.

\section{Moving Horizon Planning for Next\hyp{}Best\hyp{}Trajectories}

This section introduces the \gls*{mhp} for manipulators with $N_{\text{DoF}}$ degrees of freedom and aims at computing \glspl*{nbt}.

\paragraph*{General formulation of the \gls*{mhp}}
This paragraph provides only a brief overview of the \gls*{mhp}.
For more details, the interested reader is referred to \cite{renzMovingHorizonPlanning2024,kramerModelPredictiveControl2020,rosmannExploitingSparseStructures2018}.


The \gls*{mhp}, akin to \gls*{mpc}, iteratively optimizes a trajectory over a finite horizon. However, unlike \gls*{mpc}, which often directly applies torques to the robot, this planner solely applies joint velocities to cascaded controllers. 
The trajectory calculation is framed as a simplified \gls*{nlp} problem:
\begin{subequations}
\begin{fleqn}[0pt]
\begin{align}
    \min_{\mathbf{u}_{0:K-1}, \mathbf{x}_{1:K}}&  {J}(\mathbf{u}_{0:K-1}, \mathbf{x}_{0:K}),\\
\text{subject to:} \qquad \qquad \quad
\mathbf{x}_{k+1}  &= \mathbf{x}_k + \mathbf{u}_k \: \Delta t, \\
 \mathbf{h}(\mathbf{u}_k, \mathbf{x}_k) &\leq 0,
\end{align}
\end{fleqn}
\end{subequations}
with $k=0,1,\dots,K $. 
Within a finite horizon $K$, state vectors $\mathbf{x}_k \in \mathbb{R}^{N_{\text{DoF}}}$ and control vectors $\mathbf{u}_k\in \mathbb{R}^{N_{\text{DoF}}}$ are optimized, taking into account the cost function $J(\mathbf{u}_{0:K-1}, \mathbf{x}_{0:K})$ and the inequality constraints $ \mathbf{h}(\mathbf{u}_k, \mathbf{x}_k)$.
The sequence of states $\mathbf{x}_{0:K}$ and controls $\mathbf{u}_{0:K-1}$ is represented by state vectors $\mathbf{x}_k$ and control vectors $\mathbf{u}_k$ at each step $k$. 
$\mathbf{x}_k$ contains the robot's joint angles and $\mathbf{u}_k$ the joint velocities.
The time step $\Delta t$ is the time between two steps in a discretized time grid with $t_0 < t_1 < \ldots <t_k < \ldots < t_K$ with $k=0,1,\ldots,K$.
The cost function $J(\mathbf{u}_{0:K-1}, \mathbf{x}_{0:K})$ evaluates the trajectory's quality using a combination of quadratic terms for state $c_{\text{G}}(\mathbf{x}_k)$ and $c_{\text{C}}(\mathbf{u}_k)$ control vectors and an additional term $c_{\text{O}}(\mathbf{x}_k)$ for obstacle avoidance. 
For each step $k$, the terms are combined in $J(\mathbf{u}_k,\mathbf{x}_k) = c_{\text{G}}(\mathbf{x}_k) +  c_{\text{C}}(\mathbf{u}_k) + c_{\text{O}}(\mathbf{x}_k)$. The inequality constraints  $\mathbf{h}(\mathbf{u}_k, \mathbf{x}_k)$ encompass limits on joint angles, velocities, accelerations, and obstacle distances.
A hypergraph strategy \cite{rosmannExploitingSparseStructures2018} enhances the optimization efficiency, and \textit{IPOPT} \cite{wachterImplementationInteriorpointFilter2006} solves the \gls*{nlp}.

\paragraph*{Cost Function Extension}
The cost function $J(\mathbf{u}_{0:K-1}, \mathbf{x}_{0:K})$ is enhanced to include the \gls*{ig} throughout the trajectory, adding a cost term $ c_{\text{I}}(\mathbf{x}_{0:K}) = \sum_{k=0}^{K} \frac{w_{\text{I}}}{ O(\mathbf{x}_k) \: G(\mathbf{x}_k) + \epsilon}$.
This term is defined to account for an \gls*{of} $O(\mathbf{x}_k)$  and \gls*{ig} $G(\mathbf{x}_k)$ at each point along the trajectory, adjusted by a weighting factor $w_{\text{I}}\geq0$.
For $w_{\text{I}} = 0$, the \gls{ig} costs $c_{\text{I}}(\mathbf{x}_{0:K})$ vanishes.
The term also includes a constant $\epsilon = \num{1e-7}$ to prevent division by zero for numerical stability. 
A product instead of a sum is chosen to ensure that the costs peak if either the \gls*{of} or the \gls*{ig} is zero at state $\mathbf{x}_k$, reflecting that no gain is achieved if the orientation is not towards the \gls*{poi} or the position does not provide any \gls*{ig}.
The \gls*{ig} cost term is a barrier-like function that maximizes \gls*{ig} wherever possible. 

\paragraph*{\gls*{of} $O(\mathbf{x}_k)$ and \gls*{ig} $G(\mathbf{x}_k)$}
The \gls*{of} $O(\mathbf{x}_k)$ quantifies how well the robot's camera is aligned with the \gls*{poi} at each state $\mathbf{x}_k$.
It compares the ideal camera orientation $\mathbf{o}_{\text{POI,I}}(\mathbf{x}_k)$, towards the \gls*{poi} with the camera's actual orientation $\mathbf{o}_{\text{POI,A}}(\mathbf{x}_k)$. 
The \gls*{of} equals the cosine of the angle between the ideal and actual camera orientation and results
in $O(\mathbf{x}_k) = 1$ for perfect alignment with the target orientation and decreases for worse alignments. 
If the camera's orientation is outside the \gls*{fov}, the \gls*{of} defaults to zero, reflecting its inability to capture the \gls*{poi} within its limited viewing angle.
This metric ensures the robot's observations are as informative as possible by favoring poses where the camera is ideally oriented towards the \gls*{poi}.

The \gls*{ig} $G(\mathbf{x}_k)$ assesses the amount of environmental knowledge obtained from a specific camera position associated with the robot's joint state $\mathbf{x}_k$. 
This necessitates examining the local \gls*{id}, introduced in Section~\ref{sec:infoDist}, for each camera position throughout the trajectory. 
The local \gls*{id} is limited to $N_{\text{P}}$ perspectives, so not every camera position will coincide with these predefined points. 
To estimate the \gls*{ig} for any camera position, an adapted \gls*{idw} method interpolates the \glspl*{ig} across the nearest perspectives as
\begin{equation}
    G(\mathbf{x}_k) = \sum_{u = 0}^{N_{\text{B}}-1}\frac{1}{N_{\text{B}} - u}\frac{\sum_{j=0}^{N_{\text{P}}-1} g_{\text{P},j}  \left({d_{\text{P},j}}(\mathbf{x}_k)\right)^{-p}}{\sum_{j=0}^{N_{\text{P}}-1} \left({d_{\text{P},j}}(\mathbf{x}_k)\right)^{-p}},
\end{equation}
with the \gls*{ig} $g_{\text{P},j}$ of perspective $j$, and the distance $d_{\text{P},j}(\mathbf{x}_k)$ between the current camera position depending on $\mathbf{x}_k$ and the perspective start point of perspective $j$. 
The power $p$ serves as an adjustable parameter, allowing for the customization of the distance's impact on the interpolation process.
To ensure a balance between exploration and exploitation, $N_{\text{B}}$ previous \glspl*{id} increase the number of perspectives $N_{\text{P}}$.
The \gls{idw} result of the previous \glspl*{id} is weighted with the inverse of their position in the sequence, ensuring that older \glspl*{id} have less impact than new ones.

The combination of the \gls*{of} $O(\mathbf{x}_k)$ and the \gls*{ig} $G(\mathbf{x}_k)$ in the cost function $c_{\text{I}}(\mathbf{x}_{0:K})$ urges the robot's camera to be optimally oriented towards the \gls*{poi} while maximizing the \gls*{ig} throughout the trajectory.
A higher gain is achieved close to the sphere with radius $r_{\text{S}}$ since these perspectives cover more unknown voxels but are still close enough to the \gls*{poi} to provide valuable insights based on the choice of $r_{\text{S}}$.

\paragraph*{Global Ergodic Trajectory}
Since local planning solutions exhibit the problem of freezing robots and unknown areas due to a small horizon, a global ergodic trajectory planner \cite{dongTimeOptimalErgodic2023a} provides a reference trajectory to the local planner.
Therefore, the repository of \cite{dongTimeOptimalErgodic2023a} is extended to support manipulators.
The approach plans a trajectory that explores a given target distribution in an ergodic manner and decreases the overall motion time.
As target distribution, the discrete \gls*{id} is transferred to a continuous general mixture model using the \textit{pomegranate library} \cite{schreiberPomegranateFastFlexible}.
A fixed number $N_{\text{G}} = 3$ of Gaussian components approximates the target distribution.

The ergodic trajectory is planned in the Euclidean space and with an inverse kinematic mapped to the robot's joint space.
Note that some Euclidean points can violate the joint limits or cause self collisions, and the ergodic trajectory is then stopped at that point.
The global trajectory provides intermediate waypoints for the local planner to follow, ensuring the robot explores the environment regarding the target distribution.
Single waypoints of the global trajectory are not necessarily visited since the local planner may avoid them due to obstacles or high costs due to closeness to self collisions, joint limits, or low \gls*{ig}.
Since the runtime of the global ergodic trajectory planner is significantly higher than the local planner, the global planner is executed offline once in the beginning and can replan if the user sets new start and goal points.
The global ergodic trajectory planner is not further detailed in this paper, and the interested reader is referred to \cite{dongTimeOptimalErgodic2023a}.

\section{Experiments \& Results}
\label{sec:experiments}

The experimental setup features a \textit{Universal Robot UR10} with an \textit{Azure Kinect} RGB\hyp{}D camera mounted on the robot's end-effector for live data. 
The camera captures depth images to generate the voxel map and compute the local \gls*{id} online while the robot executes the current task. 
The voxel resolution is set at $s_{\text{Vox}} = \SI{1}{\centi\meter}$. 
The trajectory planner operates with a $t_{K} = \SI{3}{\second}$ planning horizon, a $\Delta t = \SI{0.1}{\second}$ time step, and \SI{10}{\Hz} update frequency and experiments with different information weights of $w_{\text{I}}$.
It employs an \gls*{idw} distance power of $p=2$ and an \gls{id} buffering with $N_{\text{B}}=10$. 

\subsection*{Online Capabilities of the \acrlong*{id}}
A CUDA-based implementation allows for parallel processing of \glspl*{ig} for perspectives on a \gls*{gpu}, improving efficiency and providing the planner with faster updated \gls*{ig} maps. 
This experiment examines the impact of the number of perspectives $N_{\text{P}}$ and the scaling factor $s_{\text{G}}$ on \gls*{gpu} runtimes, comparing these against a sequential \gls*{cpu} implementation. 
The comparison is conducted by replaying a recorded \textit{Azure Kinect} stream and measuring the runtime of \gls*{id} calculations on both the \gls*{cpu} and \gls*{gpu}.
The hardware comprises an \textit{Intel Core i5\hyp{}12600KF} \gls*{cpu}, \textit{NVIDIA GeForce RTX 3060} \gls*{gpu}, and \SI{32}{\giga\byte} RAM.

\begin{figure}[tb]
    \vspace{5pt} 
    \centering
    \includegraphics{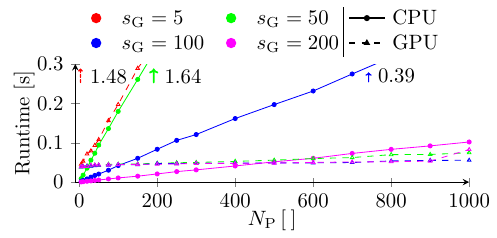}
    \captionsetup{belowskip=-15pt}
    \caption{Mean runtimes of \gls*{gpu} and \gls*{cpu} over \num{100} iterations for the \gls*{id} with various $s_{\text{G}}$ and $N_{\text{P}}$.}
    \label{fig:cpugputimes}
\end{figure}

The data in Fig.~\ref{fig:cpugputimes} illustrates the mean runtimes across \num{100} iterations for the \gls*{gpu} (dashed lines) and \gls*{cpu} (solid lines) versions of the local \gls*{id} calculation.
 The comparison covers a range of grid scaling factors $s_{\text{G}}$ and varying numbers of perspectives $N_{\text{P}}$.
To enhance clarity and emphasize the comparison, the $y$-axis in the graph is capped at \SI{0.3}{\second}. 
Note that due to the capping of the $y$-axis, the \gls*{cpu} runtime with $s_{\text{G}} = 5$ is not visible in the graph but reaches a final runtime of \SI{144,93}{\s} for $N_{\text{P}}=1000$. 
Other specific runtimes are noted directly in the figure for configurations with longer runtime. 
The \gls*{gpu} implementation outperforms the \gls*{cpu} across all configurations tested. 
Specifically, for $N_{\text{P}} = 1000$, the \gls*{gpu}'s average runtime decreases by approximately \SI{75}{\%} across various scaling factors $s_{\text{G}}$ compared to the \gls*{cpu} implementation, showcasing significant efficiency improvements.

The efficiency benefits of using the \gls*{gpu} for computations become more significant as the number of rays increases since the \gls*{gpu} implementation involves extra memory allocation and data transfer, which can be time\hyp{}consuming when dealing with a low number of rays.
This effect is clearly illustrated in Fig.~\ref{fig:cpugputimes}, where the \gls*{gpu} exhibits longer runtimes for smaller numbers of perspectives 
$N_{\text{P}}$. 
The runtime advantage gets more significant with more perspectives, for example, visible for $s_{\text{G}}=200$, where the \gls*{gpu} runtime is higher than the \gls*{cpu} runtime up to approximately $N_{\text{P}}=500$.

This study opts for $s_{\text{G}} = 100$ and $N_{\text{P}} = 500$ to maintain swift \gls*{ig} calculations for real\hyp{}time planning.

\subsection*{\acrlong*{nbt}}

This section assesses the influence of the information cost term $c_{\text{I}}$ on the \gls*{nbt} of the robot manipulator, as well as the integration of a global ergodic reference trajectory. 
It is divided into four experiments, each exploring different aspects. 
A supplementary video of the experiments is provided.
The first experiment assesses the influence of the information cost term $c_{\text{I}}$ on trajectory planning in general, while the second experiment investigates the difference compared to a baseline that applies an \gls*{nbv} strategy.
The third experiment evaluates the influence of a human working in the environment, and the fourth experiment investigates the influence of a global ergodic reference trajectory on the robot's behavior.

The evaluation metrics include the achieved product 
of \gls*{ig} and \gls*{of} and its \gls*{auc}. 
Additionally, the remaining \gls*{ig}, which reflects unseen information due to unknown voxels or revisiting previously observed voxels, is analyzed.
The reconstruction of an object at the \gls*{poi} over time is also examined by evaluating the percentage volumetric change $V_{\text{R}}$, compared to a voxel map of the object from a stationary perspective without demonstrator occlusion.
Therefore, $V_{\text{R}} = \SI{0}{\%}$ indicates a reconstruction of the reference map without additional voxels, while $V_{\text{R}} > \SI{0}{\%}$ signifies a reconstruction with more voxels, and $V_{\text{R}} < \SI{0}{\%}$ indicates a reconstruction with fewer voxels than the reference map.
It is important to note that this basic metric does not account for duplicate voxels caused by mapping inconsistencies or rapid movements and only provides an initial indication of the reconstruction potential and speed. 
This section only conducts experiments considering both \gls*{ig} and \gls*{of} terms within the costs since the combination of position and orientation is required to generate valuable views of the \gls{poi}.
\paragraph*{Experiment I}
Experiment I evaluates the impact of the information cost term $c_{\text{I}}$ on the robot's trajectory planning with different weights $w_{\text{I}}$. 
The robot task is a motion from $\mathbf{x}_{\text{start}} = \left[-1.57~0~0~0~1~0\right]^{\text{T}}$ to $\mathbf{x}_{\text{goal}} = \left[1.57~-0.75~0~0~0~0\right]^{\text{T}}$. 
While the start position is oriented towards the \gls*{poi} inside the demonstrator (see red dots Fig.~\ref{fig:teaser}), the goal position is not facing the \gls*{poi} to check the robot's capability to keep its orientation towards the \gls*{poi} during the motion.
Fig.~\ref{fig:teaser} shows the results of the robot \SI{3.5}{\s} after starting the task with different information weights $w_{\text{I}} = 0$, $w_{\text{I}} = 5$, $w_{\text{I}} = 25$, and $w_{\text{I}} = 50$ (left to right). 
It is visible that the orientation towards the \gls*{poi} benefits from increasing $w_{\text{I}}$.

\begin{figure}[tb]
    \centering
    \includegraphics{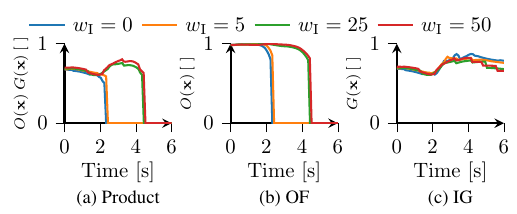}
    \captionsetup{belowskip=-15pt}
    \caption{Product (a) of \gls*{of} (b) and \gls*{ig} (c) along the \gls*{nbt} with different weights $w_{\text{I}}$ until $\mathbf{x}_{\text{goal}}$ is reached.}
    \label{fig:exp1_gain}
\end{figure}
Fig.~\ref{fig:exp1_gain} shows that increasing $w_{\text{I}}$ leads to a better orientation (b) towards the \gls*{poi} and, therefore, in an increased $O(\mathbf{x})\:G(\mathbf{x})$ (a), signifying a better data collection of the object.
Note that the time axis is capped at \SI{6}{\s} to increase the readability of the plot for lower times.
The \gls*{auc} of $O(\mathbf{x})\:G(\mathbf{x})$ over the trajectory is \num[round-mode=uncertainty,round-precision=2,separate-uncertainty]{1.34(0.03)} for $w_{\text{I}} = 0$, \num[round-mode=places,round-precision=2]{1.5295090260358104} for $w_{\text{I}} = 5$, \num[round-mode=uncertainty,round-precision=2,separate-uncertainty]{2.89(0.05)} for $w_{\text{I}} = 25$, and \num[round-mode=places,round-precision=2]{3.129601636480542} for $w_{\text{I}} = 50$. 
Note that the results of $w_{\text{I}} = 0$ and $w_{\text{I}} = 25$ are mean values and standard deviations of five trials to show that the approach is repeatable, generating similar results.
The total travel time increases with higher $w_{\text{I}}$ due to the robot's focus on areas with higher informational value, leading to an increase from \SI{7}{\s} without optimizing the \gls*{ig} to \SI{13.1}{\s} for $w_{\text{I}} = 50$.
For $w_{\text{I}} = 0$ and $w_{\text{I}} = 5$, no reconstruction of the object inside the demonstrator is possible. 
The robot can reconstruct the object for $w_{\text{I}} = 25$, and $w_{\text{I}} = 50$, and $V_{\text{R}}$ is \SI[round-mode=places,round-precision=2]{14.7246376811591}{\percent} and \SI[round-mode=places,round-precision=2]{42.7826086956562}{\percent}, respectively.
This indicates that both weights lead to a reconstruction with more voxels than the reference map due to the robot's motion.
As a compromise of the results between runtime and reconstruction, $w_{\text{I}} = 25$ is chosen for the following experiments.

\paragraph*{Experiment II}
\begin{figure}[tb]
    \centering
    \includegraphics{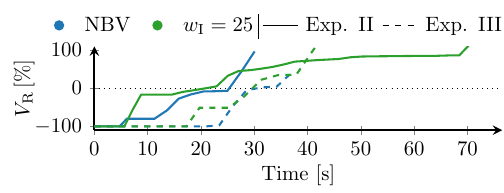}
    \captionsetup{belowskip=-15pt,aboveskip=0pt}
    \caption{$V_{\text{R}}$ over time for experiment II (solid) and III (dashed).}
    \label{fig:exp2_rekonstruktion}
\end{figure}
Experiment II compares the results of the \gls*{nbt} with a baseline that applies an \gls*{nbv} strategy \cite{santosAutonomousSceneExploration2020}. 
The robot starts at $\mathbf{x}_{\text{start}}$, and based on the \gls*{id}, the first \gls*{nbv} is selected and used for the following goal configuration. 
This is repeated for the five subsequent \glspl*{nbv} overall.
These poses serve as goals for the baseline, and the robot moves towards these without considering the optimization, $w_{\text{I}}=0$, but still using the \gls*{mhp}, and once with the \gls*{mhp} and $w_{\text{I}} = 25$.

The \gls*{auc} of the baseline is \num[round-mode=places,round-precision=2]{17.229766921928896}, and the \gls*{auc} of the \gls*{nbt} approach is \num[round-mode=places,round-precision=2]{40.25851533268002}, but the total execution time increases from \SI[round-mode=places,round-precision=2]{30.500062591}{\s} to \SI[round-mode=places,round-precision=2]{75.126779581}{\s} since the optimization of the \gls*{ig} leads to robot motions that leave areas with higher gain slower.
Due to the usage of \gls*{nbv} poses as goals, the reconstruction of the object is possible for both trials.
However, the remaining \gls*{ig} and the reconstruction time differ for the experiment.
The remaining \gls*{ig} is calculated by averaging the last $N_{\text{B}} = 10$ \gls*{ig} \glspl*{pcl}. 
For the baseline approach, the remaining \gls*{ig} is \num[round-mode=places,round-precision=2]{0.6014783302910626}; for the \gls*{nbt} approach, the remaining \gls*{ig} is \num[round-mode=places,round-precision=2]{0.46628864164054395}, signifying a reduction of \SI[round-mode=places,round-precision=2]{21,6666666666}{\%}.
The solid lines in Fig.~\ref{fig:exp2_rekonstruktion} show the reconstruction over time for both. 
It is visible that the same volume as in the reference map $V_{\text{R}}=\SI{0}{\%}$ is reached approximately \SI{4}{\s} later for the \gls*{nbv} compared to the \gls*{nbt} approach. 
Nevertheless, both approaches reconstruct the model with more voxels than the reference due to the robot motion and camera and map mismatches over time.

\paragraph*{Experiment III}
Experiment III assesses the impact of human activity within the environment on the robot's trajectory planning. 
The setup is identical to that in Experiment II, but with a human moving screws from the robot's side of the demonstrator to the opposite side, intermittently interfering with the robot's motion. 
To accommodate the added computational load, the planning horizon is shortened to $t_{K} = \SI{2}{\second}$.
The \gls*{mhp} treats the human as an obstacle and adjusts the trajectory accordingly while attempting to achieve the sequence of goals.
Note that human movements may vary between trials, leading to different outcomes for the robot, but the concept of the approach is demonstrated. 

\begin{figure}[tb]
    \vspace{5pt} 
    \centering
    \begin{subfigure}{0.48\columnwidth}
        \centering
        \includegraphics[trim=20cm 15cm 25cm 3cm,clip,width=0.8\columnwidth]{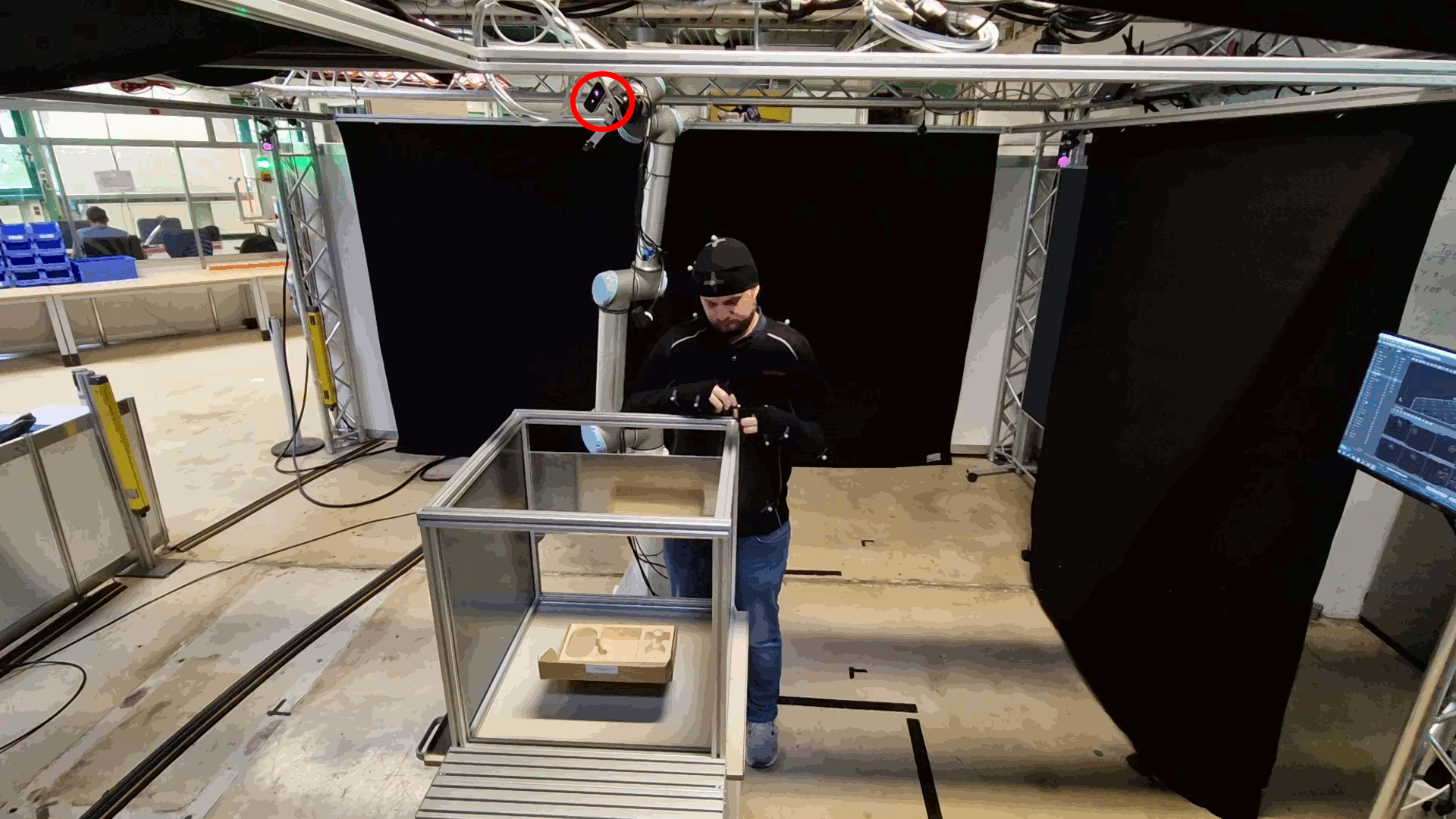}
        \captionsetup{belowskip=-5pt}
        \caption{\gls*{nbv}}
        \label{fig:nbv}
    \end{subfigure}
    \hfill
    \begin{subfigure}{0.48\columnwidth}
        \centering
        \includegraphics[trim=20cm 15cm 25cm 3cm,clip,width=0.8\columnwidth]{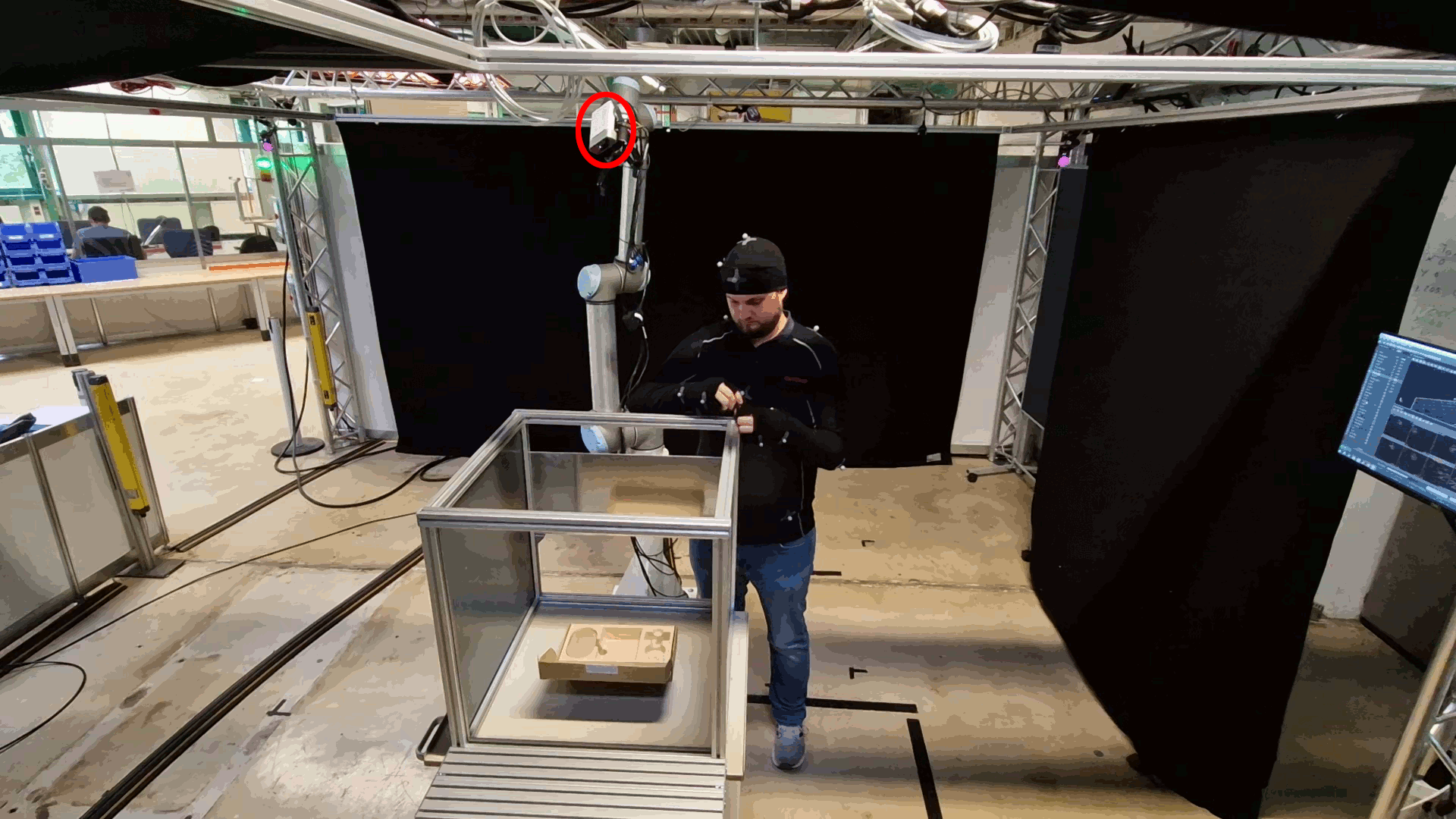}
        \captionsetup{belowskip=-5pt}
        \caption{\gls*{nbt}}
        \label{fig:nbt}
    \end{subfigure}
    \captionsetup{belowskip=-15pt}
    \caption{Working human during the robot's motion for the \gls*{nbv} (a) and \gls*{nbt} (b) approach. }
    \label{fig:exp3_human}
\end{figure}

Fig.~\ref{fig:exp3_human} directly compares \gls*{nbv} and \gls*{nbt} approaches with a human working in the environment.
It is evident that the \gls*{nbt} approach results in better orientation towards the \gls*{poi} within the demonstrator, whereas the \gls*{nbv} approach prioritizes quicker access to the first goal pose at the expense of less optimal orientation during the motion.

The dashed lines in Fig.~\ref{fig:exp2_rekonstruktion} show the reconstruction over time for \gls*{nbv} and \gls*{nbt} and that the human influences the reconstruction such that it starts later since the human blocks the way towards the first goal pose. 
However, the \gls*{nbt} reaches $V_{\text{R}}=\SI{0}{\%}$, approximately \SI{1}{\s} earlier than the \gls*{nbv} approach.

\paragraph*{Experiment IV}
Experiment IV evaluates the influence of a global ergodic reference trajectory on the robot's behavior.
Therefore, the robot starts at $\mathbf{x}_{\text{start}}$ and aims to reach $\mathbf{x}_{\text{goal}} = \left[1.57~0~0~0~0~0 \right]^{\text{T}}$.
The global ergodic reference trajectory guides the robot to the goal configuration and enables an ergodic exploration regarding the target distribution.

When relying solely on the \gls*{mhp}, the robot halts before the demonstrator and fails to reach the goal. 
In contrast, when using the ergodic reference in combination with the \gls*{mhp} with  $w_{\text{I}}=0$ and $w_{\text{I}} = 25$, the robot successfully reaches the goal pose, resulting in an increase in \gls*{ig}.
Similar to previous results, applying $w_{\text{I}} = 25$ leads to an improved view of the \gls*{poi} inside the demonstrator, increasing the \gls*{auc} by \SI[round-mode=places,round-precision=2]{28.8530202}{\%}, decreasing the remaining \gls*{ig} by \SI[round-mode=places,round-precision=2]{12.4321469}{\%} and reaching $V_{\text{R}} = \SI{0}{\%}$ approximately \SI{2}{\s} earlier.
\section{Conclusion \& Outlook}


This paper presents a novel approach to improve observation and reconstruction in robotics by maximizing \glspl*{ig} along the robot’s trajectory. 
Leveraging \gls*{gpu}-based \gls*{ig} estimation from perspectives around a \gls*{poi}, the method enables real-time \gls*{id} analysis. 
The local trajectory planner integrates this data to determine the \gls*{nbt}, optimizing for \gls*{ig} while ensuring collision avoidance, especially in human-robot interactions. 
A global ergodic planner provides a reference trajectory, enhancing exploration and preventing standstills.

Future research will refine parameter configurations for the global ergodic planner to enhance exploration and adapt \gls*{id} to different \gls*{poi} shapes. 
Additionally, integrating various \gls*{ig} metrics into the \gls*{mhp} will optimize cost functions and weighting strategies. 
Real-world experiments will examine the capability, quality and efficiency of automated data collection for training machine learning models in robotics.
An open-source OpenCL implementation of parallelized gain metric computation will be explored to benefit the broader community.




\balance
\bibliographystyle{IEEEtran}
\bibliography{literature}

\end{document}